\documentclass[11pt,a4paper]{article}
\usepackage[hyperref]{emnlp-ijcnlp-2019}
\usepackage{times}
\usepackage{latexsym}

\usepackage{url}

\aclfinalcopy 

\usepackage{array}
\newcolumntype{L}[1]{>{\raggedright\let\newline\\\arraybackslash\hspace{0pt}}m{#1}}
\newcolumntype{C}[1]{>{\centering\let\newline\\\arraybackslash\hspace{0pt}}m{#1}}
\newcolumntype{R}[1]{>{\raggedleft\let\newline\\\arraybackslash\hspace{0pt}}m{#1}}
\newcolumntype{M}[1]{>{\centering\arraybackslash}m{#1}}
\usepackage{tabularx}
\usepackage{subfigure}
\usepackage{ctable}
\usepackage{multirow}
\usepackage{amssymb}
\usepackage{bm}
\usepackage{cancel}
\usepackage{caption}
\usepackage{hyperref}
\usepackage{capt-of}

\title{Evaluating Topic Quality with Posterior Variability}

\author{Linzi Xing$^\dagger$ \and Michael J. Paul$^\ddagger$ \and Giuseppe Carenini$^\dagger$\\
  $\dagger$ University of British Columbia, Vancouver, Canada \\
  $\ddagger$ University of Colorado, Boulder, Colorado, USA \\
  {\tt \{lzxing, carenini\}@cs.ubc.ca, mpaul@colorado.edu}}

\date{}

\begin{document}
\maketitle
\begin{abstract}

Probabilistic topic models such as latent Dirichlet allocation (LDA) are popularly used with Bayesian inference methods such as Gibbs sampling to learn posterior distributions over topic model parameters.
We derive a novel measure of LDA topic quality using the {\em variability} of the posterior distributions.
Compared to several existing baselines for automatic topic evaluation,
the proposed metric achieves state-of-the-art correlations with human judgments of topic quality in experiments on three corpora.\footnote{Our code and data are available \href{https://github.com/lxing532/topic\_variability}{here}.}
We additionally demonstrate that topic quality estimation can be further improved using a supervised estimator that combines multiple metrics.

\end{abstract}

\section{Introduction}
Latent Dirichlet Allocation (LDA) \cite{blei03} topic modeling has been widely used for NLP tasks which require the extraction of latent themes, such as scientific article topic analysis \cite{Hall08}, news media tracking \cite{STM}, online campaign detection \cite{Paul14} and medical issue analysis \cite{huang-etal-2015-topic, huang2017}. To reliably utilize topic models trained for these tasks, we need to evaluate them carefully and ensure that they have as high quality as possible.  When topic models are used in an extrinsic task, like text categorization, they can be assessed by measuring how effectively they contribute to that task \cite{chen16, huang-etal-2015-topic}. However, when they are generated for human consumption, their evaluation is more challenging. In such cases, interpretability is critical, and \citet{chang09, aletras-stevenson-2013-evaluating} have shown that the standard way to evaluate the output of a probabilistic model, by  measuring  perplexity on held-out data \cite{NIPS2009}, does not imply that the inferred topics are human-interpretable.

A topic inferred by LDA is typically represented by the set of words with the highest probability given the topic. With this characteristic, we can evaluate the topic quality by determining how coherent the set of topic words is. While a variety of techniques (Section~\ref{sec:existing}) have been geared towards measuring the topic quality in this way, in this paper, we push such research one step further by making the following two contributions:  (1) We propose a novel topic quality metric by using the {\em variability} of LDA posterior distributions. This metric conforms well with human judgments and achieves the state-of-the-art performance. (2) We also create a topic quality estimator by combining two complementary classes of metrics: the metrics that use information from posterior distributions (including our new metric), along with the set of metrics that rely on topic word co-occurrence. Our novel estimator further improves the topic quality assessment on two out of the three corpora we have.

\begin{figure*}[t]
\centering 
\includegraphics[width=6.4in]{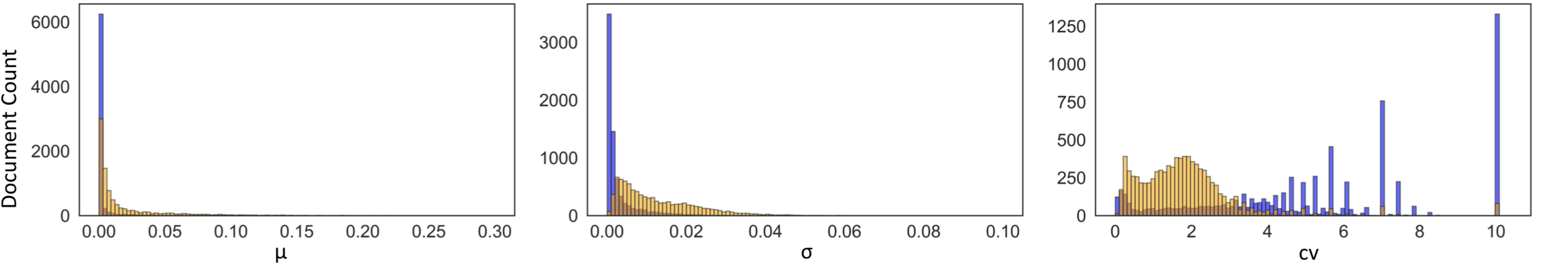}\\

\caption{\label{tab:docsamples} Two example topics and their distributions of $\mu$, $\sigma$ and $cv$ from the NYT corpus. Two topics are: \textcolor{blue}{Topic1 (in blue)}: \textit{\{financial, banks, bank, money, debt, fund, loans, investors, funds, hedge\}}. \textcolor{orange}{Topic2 (in orange)}: \textit{\{world, one, like, good, even, know, think, get, many, got\}}. Their human rating scores are 3.4 and 1.0 respectively.
}
\end{figure*}

\section{Automatic Topic Quality Evaluation}
\label{sec:existing}
There are two common ways to evaluate the quality of LDA topic models:
\textit{Co-occurrence Based Methods} and \textit{Posterior Based Methods}.
\vspace{-1ex}
\paragraph{Co-occurrence Based Methods} 
Most prominent topic quality evaluations use various pairwise co-occurrence statistics to estimate topic's semantic similarity. 
\newcite{mimno-etal-2011-optimizing} proposed the  \textbf{Coherence} metric, which is the summation of the conditional probability of each topic word given all other words. \newcite{newman-etal-2010-automatic} showed that the summation of the pairwise pointwise mutual information (\textbf{PMI}) of all possible topic word pairs is also an effective metric to assess topic quality. Later, in \citet{lau-etal-2014-machine}, PMI was replaced by the normalized pointwise mutual information (\textbf{NPMI}) \cite{Bouma2009NormalizedP}, which has an even higher  correlation with human judgments.
Another line of work exploits the co-occurrence statistics indirectly. \newcite{aletras-stevenson-2013-evaluating} devised a new method by mapping the topic words into a semantic space and then computing the pairwise distributional similarity (\textbf{DS}) of words in that space. However, the semantic space is still built on PMI or NPMI. \newcite{roder15} studied a unifying framework to explore a set of co-occurrence based topic quality measures and their parameters, 
identifying two complex combinations, (named \textbf{CV} and \textbf{CP} in that paper\footnote{The  framework proposed in \citet{roder15} has four stages. Every stage has multiple settings. CV and CP are different at the Confirmation Measure stage, which measures how strongly a set of topic words connect with each other.}), as the best performers on their test corpora.

\paragraph{Posterior Based Method} 
Recently, \newcite{xing18} analyzed how the posterior of LDA parameters vary during Gibbs sampling inference \cite{Geman84, griffins04} and proposed a new topic quality measurement named \textbf{Topic Stability}. The Gibbs sampling for LDA generates estimates for two distributions: for topics given a document ($\theta$), and for words given a topic ($\phi$). Topic stability considers $\phi$ and is defined as:

\begin{equation}
stability(\bm{\Phi_k}) = \frac{1}{|\bm{\Phi_k}|} \sum_{\phi_k \in \Phi_k} sim(\phi_{k}, \bar{\phi_k})
\end{equation}
The stability of topic $k$ is computed as the mean cosine similarity between the mean ($\bar{\phi_k}$) of all the sampled topic $k$'s distribution estimates ($\bm{\Phi_k}$) and topic $k$'s estimates from each Gibbs sampler ($\phi_{k}$). Fared against the co-occurrence based methods, topic stability is parameter-free and needs no external corpora to infer the word co-occurrence. However, due to the high frequency of common words across the corpus, low quality topics may also have high stability, and this undermines the performance of this method.

\begin{table}
\centering
\scalebox{0.9}{
\begin{tabular}{c | c c c c}

\specialrule{.1em}{.05em}{.05em}

Metric & 20NG & Wiki & NYT & Mean \\
\hline
 $\mu$ & 0.185 & 0.030 & 0.148 & 0.121 \\
 $\sigma$ & 0.480 & 0.295 & 0.600 & 0.458\\
 $cv$ & \textbf{0.679} & \textbf{0.703} & \textbf{0.774} & \textbf{0.719}\\
\specialrule{.1em}{.05em}{.05em}
\end{tabular}
}
\caption{\label{tab:res_info} Pearson's $r$ of each potential metric
of posterior variability with human judgments}
\end{table}

\begin{table*}
\centering
\scalebox{0.9}{
\begin{tabular}{p{5.8cm} | M{1.6cm} | M{1.6cm} | M{1.6cm} | M{1.6cm}}

\specialrule{.1em}{.05em}{.05em}

\textbf{Method} & \textbf{20NG} & \textbf{Wiki} & \textbf{NYT} & \textbf{Mean} \\
\hline
 CV \cite{roder15} & 0.129 & 0.385 & 0.248 & 0.254 \\
 CP \cite{roder15}  &  0.378 & 0.403 & 0.061 & 0.280\\
 DS \cite{aletras-stevenson-2013-evaluating} & 0.461 & 0.423 & 0.365 & 0.416 \\
 NPMI \cite{lau-etal-2014-machine} & 0.639 & 0.568 & 0.639 & 0.615 \\
 PMI \cite{newman-etal-2010-automatic} & 0.602 & 0.550 & 0.623 & 0.591 \\
 Coherence \cite{mimno-etal-2011-optimizing} & 0.280 & 0.102 & 0.535 & 0.305 \\
\hline
Stability \cite{xing18} & 0.230 & 0.137 & 0.322 & 0.230 \\
Variability & \textbf{0.679} & \textbf{0.703} & \textbf{0.774} & \textbf{0.719} \\
\specialrule{.1em}{.05em}{.05em}
\end{tabular}
}
\caption{\label{tab:res_individual} Pearson's $r$ correlation with human judgments for metrics.}
\end{table*}

\section{Variability Driven from Topic Estimates}
In this paper, we also use Gibbs sampling to infer the posterior distribution over LDA parameters. Yet, instead of $\phi$, our new topic evaluation method analyzes estimates of $\theta$, the topic distribution in documents.
Let $\bm{\Theta}$ be a set of different estimates of $\theta$,
which in our experiments will be a set of estimates from different iterations of Gibbs sampling.
Traditionally, the final parameter estimates are taken as the mean of all the sampled estimates, $\hat{\theta}_{dk} = \frac{1}{|\bm{\Theta}|} \sum_{\theta \in \bm{\Theta}} \theta_{dk}$. In this paper, we use the shorthand $\mu_{dk}$ to denote $\hat{\theta}_{dk}$ for a particular document $d$ and topic $k$.

In the rest of this section, we first discuss what types of information can be derived from the topic posterior estimates from different Gibbs samplers. Then, we examine how the corpus-wide variability can be effectively captured in a new metric for topic quality evaluation.

Two types of information can be derived from the topic posterior estimates: (1) the mean of estimates, $\mu_{dk}$, as discussed above, and (2) the variation of estimates. 
For variation of estimates, we considered using the standard deviation $\sigma_{dk}$. However, this measure is too sensitive to the order-of-magnitude differences of $\mu_{dk}$, that typically occur in different documents.
So, in order to capture a more stable dispersion of estimates from different iterations of Gibbs sampling, we propose to compute the variation of topic $k$'s estimates in document $d$ as its \textit{coefficient of variance (cv)} \cite{EverittBrian2002}, which is defined as: $cv_{dk} = \sigma_{dk}/\mu_{dk}$

Notice that  both $\mu_{dk}$ and $cv_{dk}$ can arguably help distinguish high and low quality topics because:
\begin{itemize}
    \item High quality topics will have high $\mu_{dk}$ for related documents and low $\mu_{dk}$ for unrelated documents. But low quality topics will have relatively close $\mu_{dk}$ throughout the corpus.
    \item High quality topics will have low $cv_{dk}$ for related documents and high $cv_{dk}$ for unrelated ones. But low quality topics will have relatively high $cv_{dk}$ throughout the corpus.
\end{itemize}

Now, focusing on the design of the new metric, we consider using the corpus-wide variability of topics' estimates as our new metric.
Figure~\ref{tab:docsamples} shows a comparison of the distributions of mean of estimates ($\mu$) and variation of estimates ($\sigma$, $cv$) for two topics across the NYT corpus (Section~\ref{sec:exp:data}). We can see the $cv$ distributions of good (Topic1) and bad (Topic2) topics are the most different. The $cv$ distribution of Topic1 covers a large span and has a heavy head and tail, while $cv$ values of Topic2 are mostly clustered in a smaller range. In contrast, the difference between Topic1 and Topic2's distributions of $\mu$ and $\sigma$ throughout the corpus appears to be less pronounced.
Taking the corpus-wide variability difference between good and bad topics observed in Figure~\ref{tab:docsamples}, we choose $cv$ to measure the corpus-wide variability of topic $k$'s estimates as our new metric. Formally, it can be defined as:
\begin{equation}
variability(k) = std(cv_{1k},cv_{2k}, \cdots ,cv_{Dk})
\end{equation} 
where $D$ is the size of the corpus. High quality topics will have higher variability and low quality topics will have lower variability. Table~\ref{tab:res_info} shows a comparison in performance  (correlation with human judgment)
of our variability defined by $cv$ with the variability defined by $\mu$ or $\sigma$ on three commonly used datasets (Section~\ref{sec:exp:data}). The variability defined by $cv$ is a clear winner. 

\section{Topic Quality Estimator}
\label{sec:estimator}

Our new method, like all other methods driven from the posterior variability, does not use any information from the topic words, which is in contrast the main driver for co-occurrence methods.
Based on this observation,  posterior variability and co-occurrence methods should be complementary to each other. To test this hypothesis, we investigate if a more reliable estimator of topic quality can be obtained by combining these two classes of methods in a supervised approach. In particular, we train a support vector regression (SVR) estimator \cite{Joachims:2006:TLS:1150402.1150429} with the estimations of these methods as features, including all the topic quality measures introduced in Section~\ref{sec:existing} along with our proposed \textit{variability} method.

\section{Experiments}
\label{sec:exp}
\subsection{Datasets}
\label{sec:exp:data}
We evaluate our topic quality estimator on three datasets: \textit{20NG}, \textit{Wiki} and \textit{NYT}. \textit{20NG} is the 20 Newsgroup dataset \cite{thor97} which consists of 9,347 paragraphs categorized into 20 classes\footnote{\url{http://archive.ics.uci.edu/ml/datasets/twenty+newsgroups}}. \textit{Wiki} consists of 10,773 Wikipedia articles written in simple English\footnote{\url{http://simple.wikipedia.org/}}. \textit{NYT} consists of 8,764 New York Times articles collected from April to July, 2016\footnote{\url{https://www.kaggle.com/nzalake52/new-york-times-articles}}.

We removed stop words, low-frequency words (appearing less than 3 times), proper nouns and digits from all the datasets, following \newcite{chang09}, so the topic modeling can reveal more general concepts across the corpus. 

Following the common setting shared by most of the papers we compared with, for each dataset we built an LDA model which consists of 100 topics represented by the 10 most probable words. The gold-standard annotation for the quality of each topic is the mean of 4-scale human rating scores from five annotators, which were collected through a crowdsourcing platform, Figure Eight\footnote{\url{https://www.figure-eight.com/}}. In order to obtain more robust estimates given the variability in human judgments, we removed ratings from annotators who failed in the test trail and recollected those with additional reliable annotators. 
To verify the validity of the collected annotations, we computed the Weighted Krippendorff's $\alpha$ \cite{krippendorff07} as the measure of Inter-Annotator Agreement (IAA) for three datasets. The average human rating score/IAA for \textit{20NG}, \textit{Wiki} and \textit{NYT} are 2.91/0.71, 3.23/0.82 and 3.06/0.69, respectively. 

\begin{table}
\centering
\scalebox{0.75}{
\begin{tabular}{c | c  c  c | c  c}
\specialrule{.1em}{.05em}{.05em}

\multicolumn{1}{c}{\textbf{Test}} & \multicolumn{3}{c}{\textbf{Train}} & \multicolumn{1}{c}{\textbf{Mean}} & \textbf{Variability}\\
\hline
\multirow{2}{*}{20NG} & Wiki & NYT & Wiki+NYT & \multirow{2}{*}{\textbf{0.801}} & \multirow{2}{*}{0.679}\\\cline{2-4}
                    & 0.790 & 0.804 & 0.810 \\
\hline
\multirow{2}{*}{Wiki} & 20NG & NYT & 20NG+NYT & \multirow{2}{*}{\textbf{0.716}} & \multirow{2}{*}{0.703}\\\cline{2-4}
                                 & 0.707 & 0.731 & 0.710 \\
\hline
\multirow{2}{*}{NYT} & 20NG & Wiki & 20NG+Wiki & \multirow{2}{*}{0.770} & \multirow{2}{*}{\textbf{0.774}}\\\cline{2-4}
                                 & 0.762 & 0.775 & 0.773 \\
\specialrule{.1em}{.05em}{.05em}
\end{tabular}
}
\caption{\label{tab:res_estimator} Pearson's $r$ correlation with human judgments for the topic quality estimator.}
\end{table}

\subsection{Experimental Design}
\label{sec:exp:design}
\paragraph{Topic Modeling} Following the settings in \citet{xing18}, we ran the LDA Gibbs samplers for 2,000 iterations \cite{griffins04} for each datasets, with 1,000 burn-in iterations, collecting samples every 10 iterations for the final 1,000 iterations. The set of estimates $\bm{\Theta}$ thus contains 100 samples.
\paragraph{Estimator Training}
was performed following the cross-domain training strategy \cite{bhatia-etal-2018-topic}. With the ground truth (human judgments), we train the estimator on all topics over one dataset, and test it on another (\textbf{one-to-one}). To enlarge the training set, we also train the estimator on two datasets merged together and test it on the third one (\textbf{two-to-one}). Given the limited amount of data and the need for interpretability, we experimented only with non-neural classifiers, including linear regression, nearest neighbors regression, Bayesian regression, and Support Vector Regression (SVR) using \texttt{sklearn} \cite{scikit-learn}; we report the results with SVR, which gave the best performance. We also experimented with different kernels of SVR and rbf kernel worked best.
\paragraph{Baselines} include seven commonly adopted measures for topic quality assessment: \textbf{CV}, \textbf{CP} \cite{roder15}, \textbf{DS} \cite{aletras-stevenson-2013-evaluating}, \textbf{PMI} \cite{newman-etal-2010-automatic}, \textbf{NPMI} \cite{mimno-etal-2011-optimizing}, \textbf{Coherence} \cite{newman-etal-2010-automatic} and \textbf{Stability} \cite{xing18}. All of them are introduced in Section~\ref{sec:existing}.

\begin{figure}
\centering
\includegraphics[width=\columnwidth, clip]{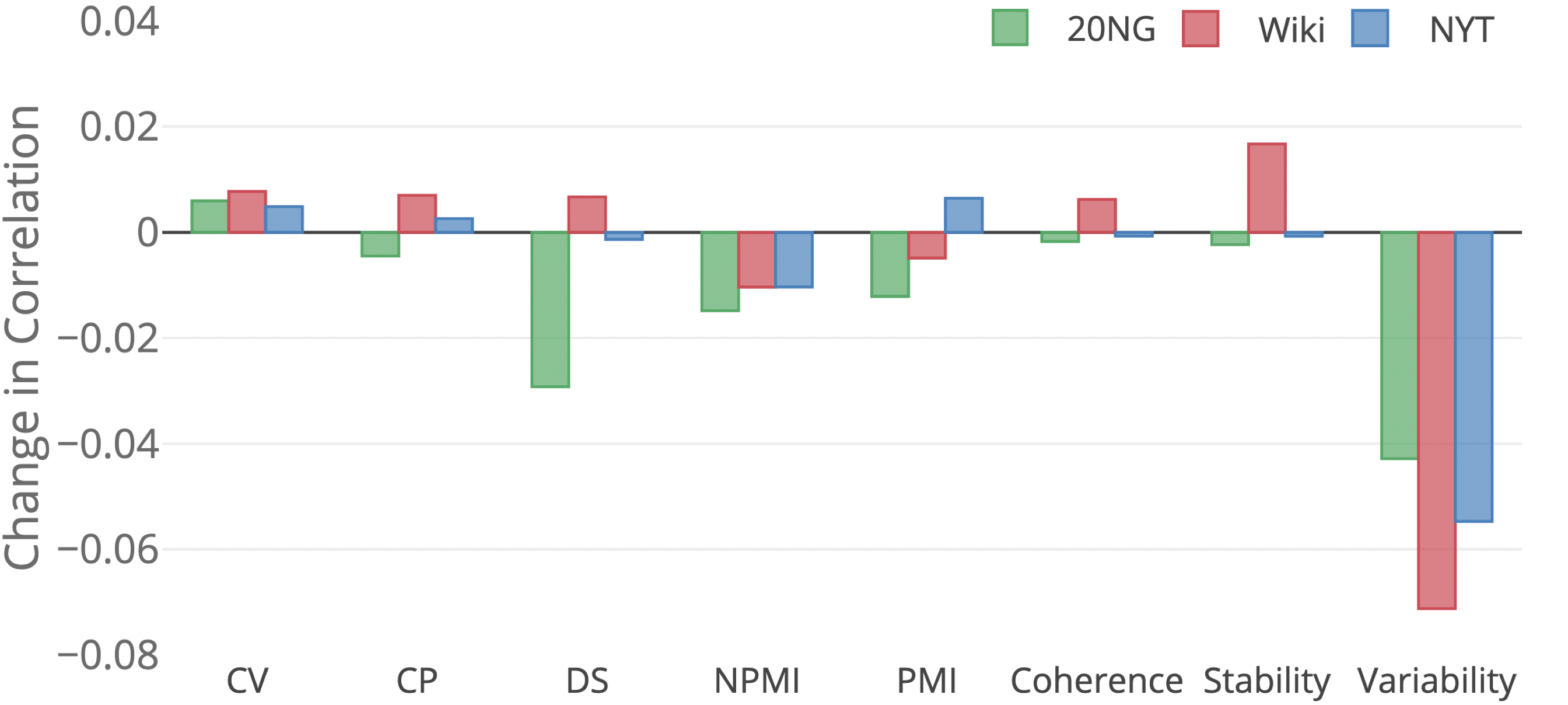}
\caption{\label{fig:ablation} 
The ablation analysis.}
\end{figure}

\begin{center}
\begin{figure*}
\centering
\begin{minipage}[t]{1.0\textwidth}
\centering
\includegraphics[width=5.4in]{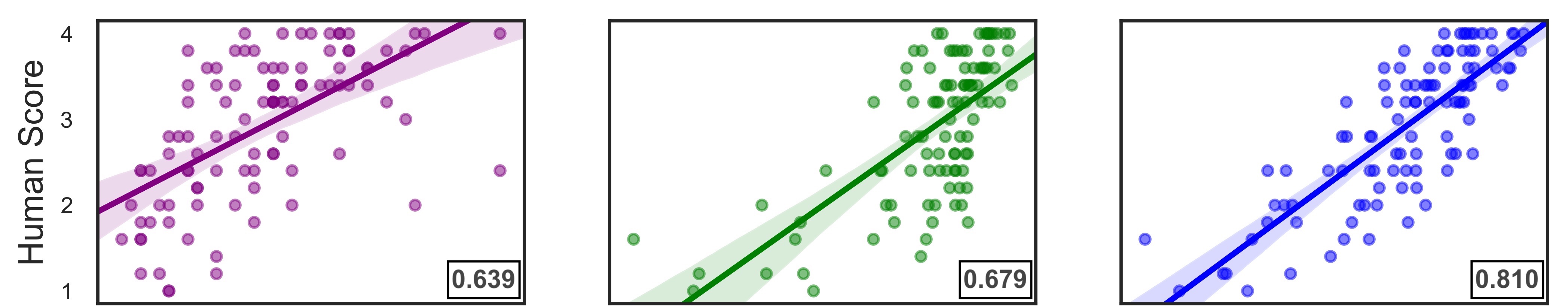}
\end{minipage}%
\\
\begin{minipage}[t]{1.0\textwidth}
\centering
\includegraphics[width=5.4in]{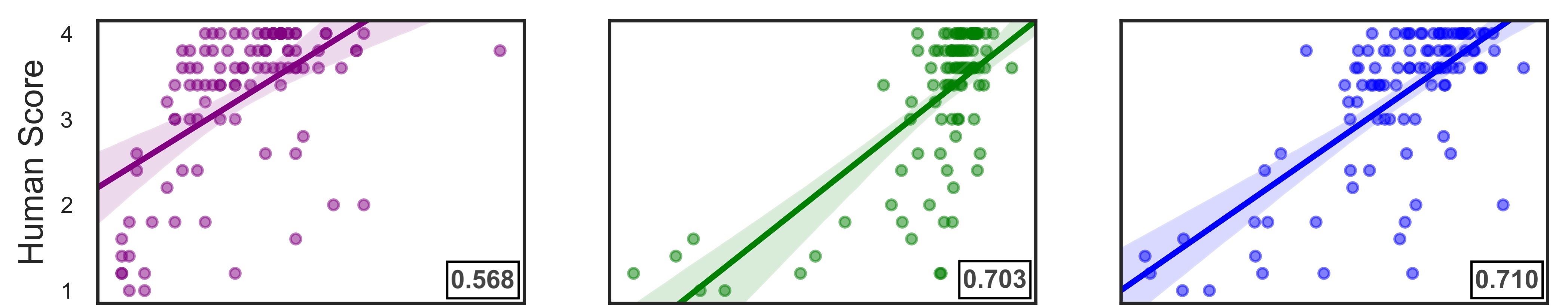}
\end{minipage}%
\\
\begin{minipage}[t]{1.0\textwidth}
\centering
\includegraphics[width=5.4in]{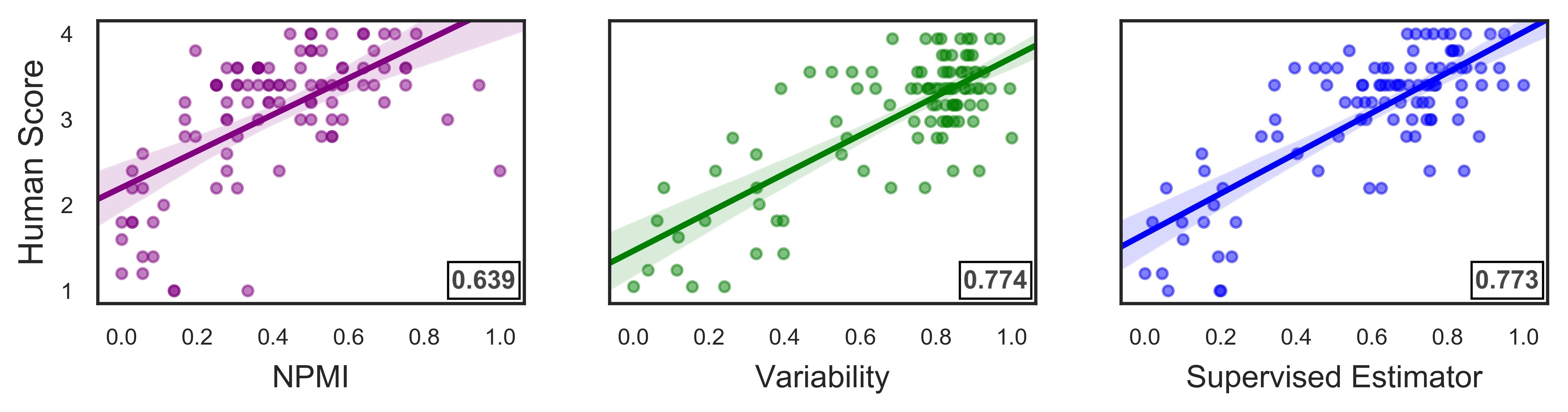}
\end{minipage}%
\caption{\label{fig:scatter} 
Scatter plots illustrating the correlation between human rating scores and the three metrics: \textit{NPMI}, \textit{Variability} and \textit{Supervised Estimator} on three datasets: \textit{20NG} (top row), \textit{Wiki} (middle row) and \textit{NYT} (bottom row). The numerical Pearson's $r$ correlations are  shown in the bottom-right corner. }
\end{figure*}
\end{center}

\vspace{-5ex}
\subsection{Results}
\label{sec:exp:res}
Following \cite{roder15}, we use Pearson's $r$ to evaluate the correlation between the human judgments and the topic quality scores predicted by all the automatic metrics. The higher is the Pearson's $r$, the better the metric is at measuring topic quality. Table~\ref{tab:res_individual} shows the Pearson's $r$ correlation with human judgments for all the metrics. Our proposed variability-based metric substantially outperforms all the baselines.

Table~\ref{tab:res_estimator} shows the Pearson's $r$ correlation with our proposed topic quality estimator trained and tested on different datasets. The average correlations of the estimator dominates our proposed variability-based metric on two out of three datasets, and on one of them by a wide margin.

Additionally, to better investigate how well the metrics align with human judgments, in Figure~\ref{fig:scatter} we use scatter plots to visualize their correlations and make the following observations. 
The top performer co-occurrence based metric, \textit{NPMI},  tends to underestimate topic quality by giving low ratings to relatively high-quality topics (dots with high human scores tend to be above the purple line), but it performs relatively well for low-quality topics. On the contrary, the top performer posterior based metric, \textit{variability}, is more likely to overestimate topic quality by assigning high ratings to relatively bad topics (dots with low human scores tend to be below the green line), but it performs relatively well for high-quality topics. Thus, when we combine all the metrics in a supervised way, the topic quality estimation becomes more accurate, especially on 20NG corpus (i.e. the top row).

\vspace{-1ex}
\paragraph{Ablation Analysis:} Since some features in the topic quality estimator are closely related, their overlap/redundancy may even hurt the model's performance.
To better understand the contributions of each feature in our proposed estimator, we conduct ablation analysis whose results are illustrated in Figure~\ref{fig:ablation}. We track the change of performance by removing one feature each time. The more significant drop in performance indicates that the removed feature more strongly contributes to the estimator's accuracy. By training on two datasets and testing on the third dataset, we find that only \textit{Variability} and \textit{NPMI} consistently contributes to accurate predictions on all three datasets. This indicates that our new \textit{Variability} metric and \textit{NPMI} are the strongest ones from the two families of Posterior-based and Co-occurrence-based metrics, respectively.

\section{Conclusion and Future Work}
We propose a novel approach to estimate topic quality grounded on the variability of the variance of LDA posterior estimates. We observe that our new metric, driven by Gibbs sampling, is more accurate than previous methods when tested against human topic quality judgment. Additionally, we propose a supervised topic quality estimator that by combining multiple metrics delivers even better results. For future work, we intend to work with larger datasets to investigate neural solutions to combine features from different metrics, as well as to apply our findings to other variants of LDA models trained on low-resource languages, where high-quality external corpora are usually not available \cite{hao-etal-2018-lessons}.

\section*{Acknowledgments}
We would like to thank the anonymous reviewers for their valuable suggestions and comments, as well as Jey Han Lau and Data Science Group at Universitat Paderborn for making their topic quality evaluation toolkit publicly available.

\bibliographystyle{acl_natbib}
\bibliography{emnlp-ijcnlp-2019}

\appendix

\section{Guidelines for Human Annotation}

\begin{figure*}
    \centering
    \includegraphics[height=0.8\vsize, width=0.82\hsize]{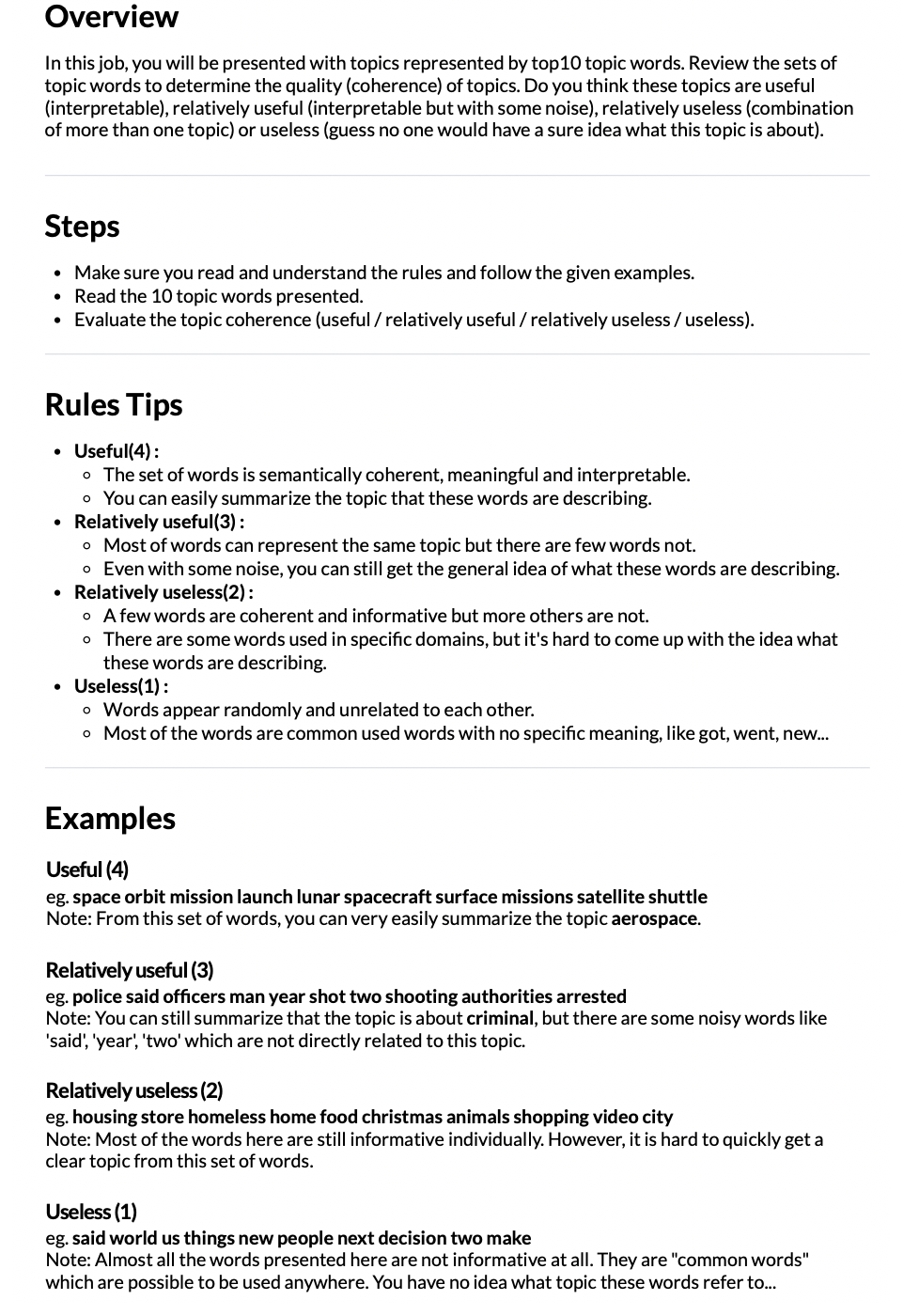}
    \captionof{figure}{The details of the guidelines we provided to the topic quality annotators. All topics were rated on a 4-point Likert scale. In particular, we provided the descriptions of the meaning of the four criteria, as well as the made-up examples (one for each score) that were created to help raters understand the criteria.}
\end{figure*}

\end{document}